\documentclass[letterpaper]{article} 
\usepackage{aaai24}  
\usepackage{times}  
\usepackage{helvet}  
\usepackage{courier}  
\usepackage[hyphens]{url}  
\usepackage{graphicx} 
\usepackage{natbib}  
\usepackage{caption} 
\usepackage{multirow}
\usepackage[ruled,linesnumbered]{algorithm2e}
\usepackage{algpseudocode}
\usepackage{subfigure}
\usepackage{xcolor}
\usepackage{color}
\usepackage[toc,title]{appendix}
\usepackage{enumitem}
\newcommand{\modelname}{PoinDP}
\usepackage{float}
\usepackage{amsmath}
\usepackage{amsthm}
\usepackage{booktabs}
\usepackage{amssymb}
\usepackage{adjustbox}
\usepackage{makecell}

\newtheorem{myDef}{Definition}
\newtheorem{thm}{\bf Theorem}

\usepackage{newfloat}
\usepackage{listings}
\DeclareCaptionStyle{ruled}{labelfont=normalfont,labelsep=colon,strut=off} 
\floatstyle{ruled}
\newfloat{listing}{tb}{lst}{}
\floatname{listing}{Listing}
\title{Poincaré Differential Privacy for Hierarchy-Aware Graph Embedding}
\author{
    {Yuecen Wei\textsuperscript{\rm 1,\rm 2,\rm 3}, Haonan Yuan\textsuperscript{\rm 1}, Xingcheng Fu\textsuperscript{\rm 3}\thanks{Corresponding authors}, Qingyun Sun\textsuperscript{\rm 1}, Hao Peng\textsuperscript{\rm 1}, \\
    Xianxian Li\textsuperscript{\rm 3}, Chunming Hu\textsuperscript{\rm 1,\rm 2}\footnotemark[1]} \\
}
\affiliations{
    \textsuperscript{\rm 1}Beijing Advanced Innovation Center for Big Data and Brain Computing, Beihang University, Beijing, China\\
    \textsuperscript{\rm 2}School of Software, Beihang University, Beijing, China\\
    \textsuperscript{\rm 3}Key Lab of Education Blockchain and Intelligent Technology, Ministry of Education, Guangxi Normal University, China\\  
    \{weiyc,yuanhn,sunqy,penghao,hucm\}@buaa.edu.cn, 
    \{fuxc,lixx\}@gxnu.edu.cn \\
}

\usepackage{bibentry}

\begin{document}

\maketitle

\begin{abstract}
Hierarchy is an important and commonly observed topological property in real-world graphs that indicate the relationships between supervisors and subordinates or the organizational behavior of human groups. 
As hierarchy is introduced as a new inductive bias into the Graph Neural Networks (GNNs) in various tasks, it implies latent topological relations for attackers to improve their inference attack performance, leading to serious privacy leakage issues. 
In addition, existing privacy-preserving frameworks suffer from reduced protection ability in hierarchical propagation due to the deficiency of adaptive upper-bound estimation of the hierarchical perturbation boundary. 
It is of great urgency to effectively leverage the hierarchical property of data while satisfying privacy guarantees. 
To solve the problem, we propose the \textbf{Poin}car\'e \textbf{D}ifferential \textbf{P}rivacy framework, named \textbf{PoinDP}, to protect the hierarchy-aware graph embedding based on hyperbolic geometry. 
Specifically, PoinDP first learns the hierarchy weights for each entity based on the Poincar\'e model in hyperbolic space. 
Then, the Personalized Hierarchy-aware Sensitivity
is designed to measure the sensitivity of the hierarchical structure and adaptively allocate the privacy protection strength.
Besides, the Hyperbolic Gaussian Mechanism (HGM) is proposed to extend the Gaussian mechanism in Euclidean space to hyperbolic space to realize random perturbations that satisfy differential privacy under the hyperbolic space metric. 
Extensive experiment results on five real-world datasets demonstrate the proposed PoinDP's advantages of effective privacy protection while maintaining good performance on the node classification task. 
\end{abstract}

\section{Introduction}
\label{sec:intro}
The inherent topological properties of graphs have been widely leveraged in graph representation learning as inductive biases~\cite{sunqy2022IB, AGE2023TKDErobust, zhang2024learning, yuan2023environment}. 
Real-world graph data typically exhibit intricate topological structures with diverse properties~\cite{sunqyWWW21SUGAR}, and the \textbf{hierarchy} frequently assumes a pivotal role, which naturally mirrors human behavior within hierarchical organizations. 
This property assists in the learning of graph representation by capturing implicit data organization patterns~\cite{papadopoulos2012popularity}. 
However, dealing directly with hierarchy in the topological space proves to be challenging in the Euclidean embedding space. 
In contrast to the Euclidean space, the hyperbolic geometric space can be conceptualized as a continuous tree-like structure, naturally capable of representing the topological hierarchy~\cite{Sunli23RiemanTempGL, Sunli23AdapRieman}. 
The constant negative curvature of the hyperbolic geometric space imparts it with a more potent ability for hierarchical representation compared to the flat Euclidean space~\cite{Krioukov2010Hyperbolic, Sunli23CONGREGATE}. 
Recent works in hyperbolic representation learning~\cite{HNN:GaneaBH18, PoincareGlove} have achieved noteworthy success by harnessing the hierarchy, where the hierarchy is regarded as the prompt for balancing the aggregation weights. 
These methods manifest how the hierarchy can be utilized to enhance the effectiveness of graph representation learning. 

\begin{figure}[t]
    \begin{minipage}[b]{.45\linewidth}
    \centering
    \subfigure[Latent Tree Structure.]{
    \includegraphics[width=1\linewidth, trim=330 10 240 5, clip]{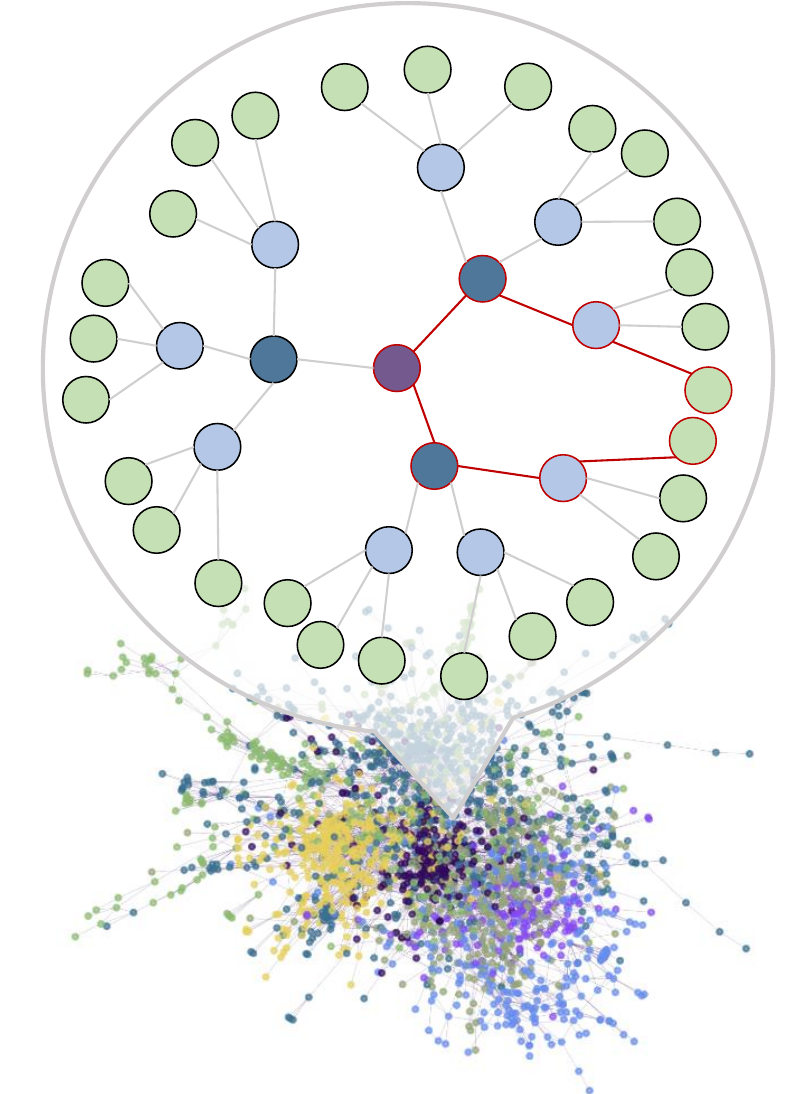}
    }%
    \end{minipage}
    \hspace{0.2cm}
    \begin{minipage}[b]{.5\linewidth}
        \centering
        \subfigure[Neighbour Classification.]{
        \includegraphics[width=1\linewidth]{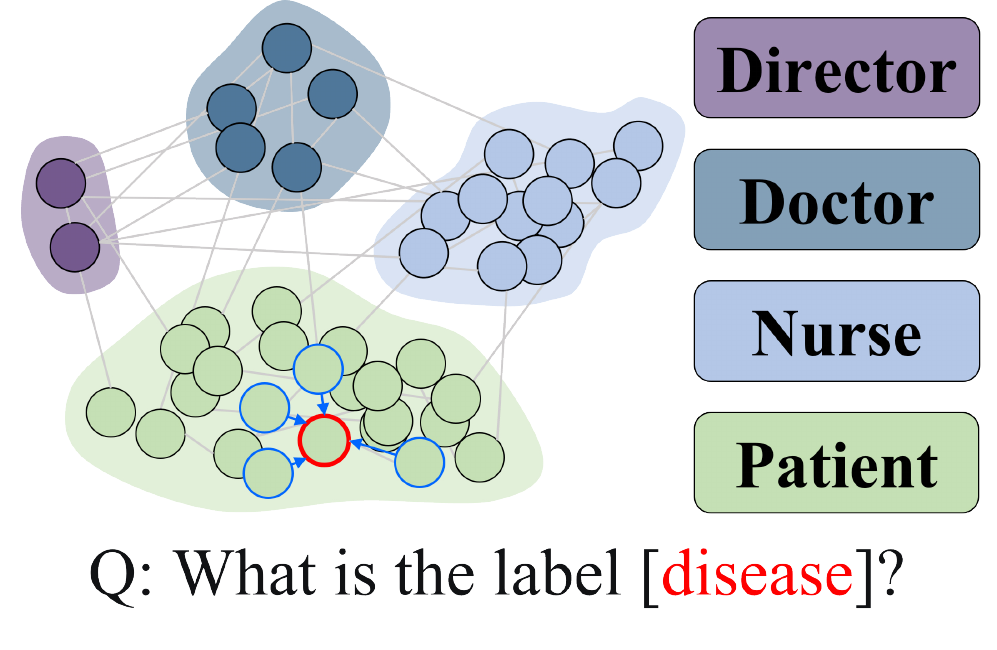}
        }
        \subfigure[Hierarchical Classification.]{
        \includegraphics[width=1\linewidth]{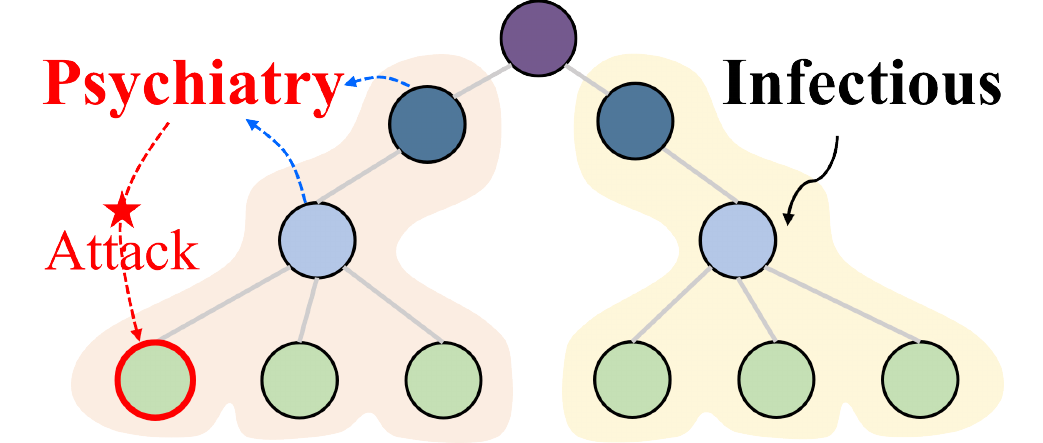}
        }%
    \end{minipage}
\caption{Privacy leakage on the hierarchical structure. }
\label{fig:dpexample}
\end{figure}

However, while representation learning in the hyperbolic geometric space offers benefits, it comes with the potential drawback of increased susceptibility to the leakage of sensitive user information. 
Hierarchical structures are prevalent in graph data, as depicted in Figure~\ref{fig:dpexample}(a). 
Euclidean and hyperbolic spaces exhibit distinct perceptual capabilities for hierarchical structures. 
For instance, in Euclidean space, clustering based on node distances can reflect whether an individual is a patient, while node distances in hyperbolic space can indicate the type of ailment.  
In addition, traditional Euclidean Graph Neural Networks (GNNs) primarily focus on neighborhood aggregation and struggle to capture latent hierarchical tree-like structures, as illustrated in Figure~\ref{fig:dpexample}(b). 
Consequently, the feature perturbation conducted by conventional GNN privacy frameworks only disturbs connections among nodes of the same category, to thwart attacker inferences. 
However, \textbf{although hyperbolic GNNs provide a direct approach to learning hierarchical structural features, they concurrently elevate the risk of privacy leakage}, as depicted in Figure~\ref{fig:dpexample}(c). 
Attackers, without accessing sensitive information, can deduce patients' medical conditions. 
For example, within the hierarchical structure, an attacker can infer that patients affiliated with psychiatric departments are likely to have mental illnesses, and patients in infectious disease departments have contagious diseases. 

To address privacy concerns, notable privacy-preserving techniques have been proposed. 
Differential privacy (DP)~\cite{DP2006, DP20062} stands out as one of the most prominent methods due to its robust mathematical foundation. 
However, existing DP methods tailored for graph representation learning have predominantly centered on safeguarding node features and neighborhood structures~\cite{pengCIKM2022, DPGGAN2021,HeteDP}, with a limited focus on preserving implicit topological properties such as hierarchy. 
This underscores the need for novel strategies that holistically address both the intricate topological features and privacy considerations within graph representation learning. 

To utilize the geometric prior of the hyperbolic space to capture the hierarchy properties and guarantee that the sensitive information in the hierarchy, the major problems are as follows: 
$(1)$ Traditional privacy-preserving methods usually consider the privacy between neighbors or relations to generate perturbation noise, which is weak to capture the hierarchical structure of the graph. 
$(2)$ Existing privacy-preserving techniques measure the privacy of nodes in Euclidean space, which doesn't work in hyperbolic space due to the Gaussian mechanism based on the standard normal distribution just defined in Euclidean space. 

\textbf{Present work.} To address the above problems, we propose a novel \textbf{Poin}car\'e \textbf{D}ifferential \textbf{P}rivacy framework for protecting hierarchy-aware graph embedding based on hyperbolic geometry, named \textbf{PoinDP}\footnote{Code is available at https://github.com/WYLucency/PoinDP.}. 
First, the Personalized Hierarchy-aware Sensitivity (PHS) is designed to utilize the Poincar\'e model to capture the inter- and intra-hierarchy node information. 
PHS can allocate the privacy budget between \textit{radius} (inter-hierarchy) and \textit{angle} (intra-hierarchy) and learn high-quality graph representations effectively while satisfying the differential privacy guarantee. 
Then, a novel Hyperbolic Gaussian Mechanism (HGM) extends the Gaussian mechanism in Euclidean space to hyperbolic space to realize random perturbations that satisfy differential privacy under the hyperbolic space metric for the first time. 
Extensive experimental results conducted on five datasets empirically demonstrate that \modelname~has consistent advantages. 
We summarize our contributions as follows: 
\begin{itemize}[leftmargin=*]
    \item We propose a novel Poincar\'e differential privacy for hierarchy-aware graph embedding framework named~\modelname. To the best of our knowledge, this is the first work that presents the privacy leakage problem due to the hierarchical structure and gives a definition of the privacy problem in terms of hyperbolic geometry. 
    \item In \modelname, the Personalized Hierarchy-aware Sensitivity can measure the sensitivity of the hierarchical structure and adaptively allocate the privacy protection strength. 
    Besides, we extend the Gaussian mechanism in Euclidean space to hyperbolic space to realize random perturbations that satisfy differential privacy under the hyperbolic space metric for the first time, which can be used in other hyperbolic privacy works to promote community development. 
    \item Experiments demonstrate that~\modelname~can effectively resist attackers with hierarchical information enhancement, and learn high-quality graph representations while satisfying privacy guarantees. 
\end{itemize}

\section{Related Work}

\subsection{Graph Neural Networks} 

In the field of graph representation learning, Graph Neural Networks (GNNs) have achieved remarkable success in learning embeddings from graph-structured data due to their powerful graph representation capabilities, while are widely extended for downstream tasks in complex scenarios~\cite{GCN, GAT, Graphsgae}. 
However, traditional GNNs operating in Euclidean space often fall short of effectively utilizing the topology properties of graphs, leading to suboptimal semantic understanding, particularly overlooking the hierarchical relationships within the data, which is of vital importance in real-world scenarios. 

Recently, certain categories of data (e.g., hierarchical, scale-free, or spherical data) have demonstrated superior representation capabilities when modeled through non-Euclidean geometries. This has led to a burgeoning body of work on deep learning~\cite{PoincareGlove,sala2018representation, HNN:GaneaBH18}. Notably, hyperbolic geometric spaces have garnered significant attention and adoption within the domain of graph representation learning~\cite{HGNN_Qi, HGCN_ChamiYRL19, bachmann2020constant, sun2021hyperbolic, HyperIMBA, sunqyCIKM2022Imbalance, wu2022}, attributed to their inherent capacity and prowess in preserving hierarchical structures~\cite{Sunli23MixedCurv}. 

However, with the evolution of increasingly intricate models aimed at extracting potential correlations among nodes, 
the complex structure inadvertently amplifies the attackers' capacity for inference, enabling lateral enhancement of their inferential ability. 
Unfortunately, the majority of GNN-based methodologies have been demonstrated to possess vulnerabilities susceptible to inference attacks~\cite{MIA3_TPS-ISA2021, MIA4_USENIXSecurity2022}.

\subsection{Differentially Private GNNs} 
Differential privacy (DP) ~\cite{DP2006} is a privacy protection method and introduces random noise perturbation mechanisms to the original data, ensuring that attackers cannot infer the original data from the outputs of models. 
For graph privacy protection, we divide the existing DP method into two levels: node-level and edge-level. 

For node-level DP, the works focus on perturbing node features or node labels to execute privacy protection. 
AsgLDP ~\cite{asgldpTIFS2020} proposed randomized attribute lists (RAL) to perturb each bit of node feature by the randomized response, and LPGNN ~\cite{lpgnnCCS2017} used a multi-bit mechanism to sample perturbed features while using the randomized response to mask node labels. 
GAP ~\cite{gapUSENIX_Security2023} perturbed the out of each aggregation using Gaussian noise while saving them.  
HeteDP~\cite{HeteDP} utilized meta-path to adapt data heterogeneity while personalized node perturbation by multi-attention mechanism. 

For edge-level DP, the target of noise addition is the topology of the graph, e.g. the degree and the adjacency matrix that represent the information about the interactions between nodes. 
LDPGEN ~\cite{ldpgenCCS2017} computed each subgroup degree vector on the client, then uploaded it to the server and exerted Laplace noise to the degree vectors. The graph structure generation uses the BTER model. 
Solitude ~\cite{solitudeTIFS2022} used the randomized response to flip graph adjacency matrix and a regularization term to optimize noise. 
LF-GDPR ~\cite{lf-gdprTKDE2022} perturbed node degrees and adjacency matrix in the client and the server will receive a double-degree message to aggregate and calibrate. 

However, most DP schemes are deficient in adaptability to complex structures and hardly explore potential properties to adjust perturb design.

\section{Preliminary}
\label{sec:Problem Definition}

Differential privacy ~\cite{DP2006} is considered to be one of the quantifiable and practical privacy-preserving data processing techniques. 
It protects privacy by adding noise to the query results, and an attacker cannot infer any information from these query results even if he or she knows all the records except this particular individual information. 

\begin{myDef}[$(\epsilon, \delta)$-Differential Privacy]
	\label{def:dp}
    Given two adjacent datasets $\mathcal{D} $ and ${\mathcal{D}}'$ differ by at most one record, and they are protected via a random algorithm $\mathcal{M} $, which satisfies $\left ( \epsilon, \delta  \right ) $-differential privacy (DP)~\cite{DP20062}. For any possible subset of output 
    $\mathcal{O} \subseteq \mathrm{Range}\left ( \mathcal{M}\right ) $, we have 
    \begin{equation}
    \begin{aligned}
        \mathrm{Pr} \left [ \mathcal{M} \left ( \mathcal{D} \right ) \in \mathcal{O} \right ] \le \mathrm{e}^{\epsilon } \mathrm{Pr} \left [\mathcal{M} \left ( {\mathcal{D}}'  \right ) \in \mathcal{O} \right ] +  \delta , 
    \end{aligned}
    \end{equation}
    where $\epsilon$ is the privacy budget, $\delta$ is a probability to break $\epsilon$-DP and $\mathrm{Range}( \mathcal{M})$ denotes the value range of $\mathcal{M}$ output. 
\end{myDef}

\begin{myDef}[Sensitivity]
Given any query $\mathcal{S} $ on $D$, the sensitivity~\cite{DP20062} for any neighboring datasets $D$ and $D^{\prime }$ are defined as
\begin{equation}\label{eq:defineSens}
\begin{aligned}
    \Delta _{2} \mathcal{S} =\max_{D,D^{\prime } } \left \| \mathcal{S} \left ( D \right ) - \mathcal{S} \left ( D^{\prime }  \right )   \right \|_{2} .
\end{aligned}
\end{equation}
\end{myDef}

\begin{myDef}[Gaussian Mechanism]
Let $\mathcal{S} : D \to \mathbf{O} ^{\mathcal{K} } $ be an arbitrary $\mathcal{K}$-dimensional function and define its $L_{2}$ sensitivity to be $\Delta _{2} \mathcal{S}$. 
The Gaussian Mechanism~\cite{GaussianDP2014} with parameter $\sigma $ adds noise scaled to $\mathcal{N} \left (0,\sigma ^{2}  \right ) $ to each of the $n$ components of the output. 
Given $\epsilon \in \left ( 0,1 \right ) $, the Gaussian Mechanism is $\left ( \epsilon ,\delta  \right ) $-DP with
\begin{equation}
\begin{aligned}
    \sigma \ge \sqrt{2\ln_{}{\left ( 1.25/\delta  \right ) } } \Delta _{2} \mathcal{S}/\epsilon .
\end{aligned}
\end{equation}
\end{myDef}
For a graph $\mathcal{G}$, the overall form of the perturbed noise is defined as
\begin{equation}\label{eq:GaussionNoise}
\begin{aligned}
    \mathcal{M} \left ( \mathcal{G} \right ) \overset{\triangle }{=} \mathcal{S} \left ( \mathcal{G} \right ) +\mathcal{N} \left ( 0,\left ( \triangle _{2}\mathcal{S} \right )^2 \sigma^2   \right ) ,
\end{aligned}
\end{equation}
where $\Delta _{2} \mathcal{S}$ controls the amount of noise in the generated Gaussian distribution from which we will sample noise into the target. 

Our goal is to keep $(\epsilon, \delta)$-DP effective in high-dimensional projection spaces and message passing while maintaining classification performance. 
Compared to the traditional Euclidean space, hyperbolic space has a stronger hierarchical structure. 
The Poincaré ball model~\cite{NickelK17Poincare} is a commonly used isometric model in hyperbolic space, and we exploit it to capture the latent hierarchical structure of the graph.
\begin{myDef}[Poincar\'e Ball Model]
	\label{def:poincare}
    Given a constant negative curvature $c$, Poincar\'e Ball $\mathcal{B} ^{n}$ is a Riemannian manifold $(\mathcal{B} _{c}^{n}, g_{x}^{\mathcal{B}} )$, where $\mathcal{B} _{c}^{n}$ is an $n$-dimensional ball of radius $1/\sqrt{c}$ and $g_{x}^{\mathcal{B}}$ is metric tensor. The Poincar\'e distance between the node pair $(\mathbf{x},\mathbf{y})$ is defined as
    \begin{equation}\label{Eq:hyperbolic_distance}
        d_{\mathcal{B}^{n}_{c}}(\mathbf{x}, \mathbf{y}) = \frac{2}{\sqrt{c}} \tanh^{-1} (\sqrt{c} \left\| -\mathbf{x} \oplus_c \mathbf{y} \right\|),
    \end{equation}
    where $\oplus_c$ is M\"{o}bius addition and $\left\| \cdot \right\|$ is $L_2$ norm.
\end{myDef}

\begin{myDef}[Poincar\'e Norm]\label{def:Poincare_norm}
The Poincar{\'{e}} Norm is defined as the distance of any point $\mathbf{x} \in \mathcal{B}^{n}_{c}$ from the origin of Poincar{\'{e}} ball:
\begin{equation}
\label{Eq:distortion}
\mathrm{Norm}_{\mathcal{B}^{n}_{c}} (\mathbf{x}) = \left\| \mathbf{x} \right\|_{\mathcal{B}^{n}_{c}} = \frac{2}{\sqrt{c}} \tanh^{-1} (\sqrt{c} \left\| \mathbf{x} \right\|).
\end{equation}
\end{myDef}

\begin{figure*}[ht]
    \centering
    \includegraphics[width=1\linewidth]{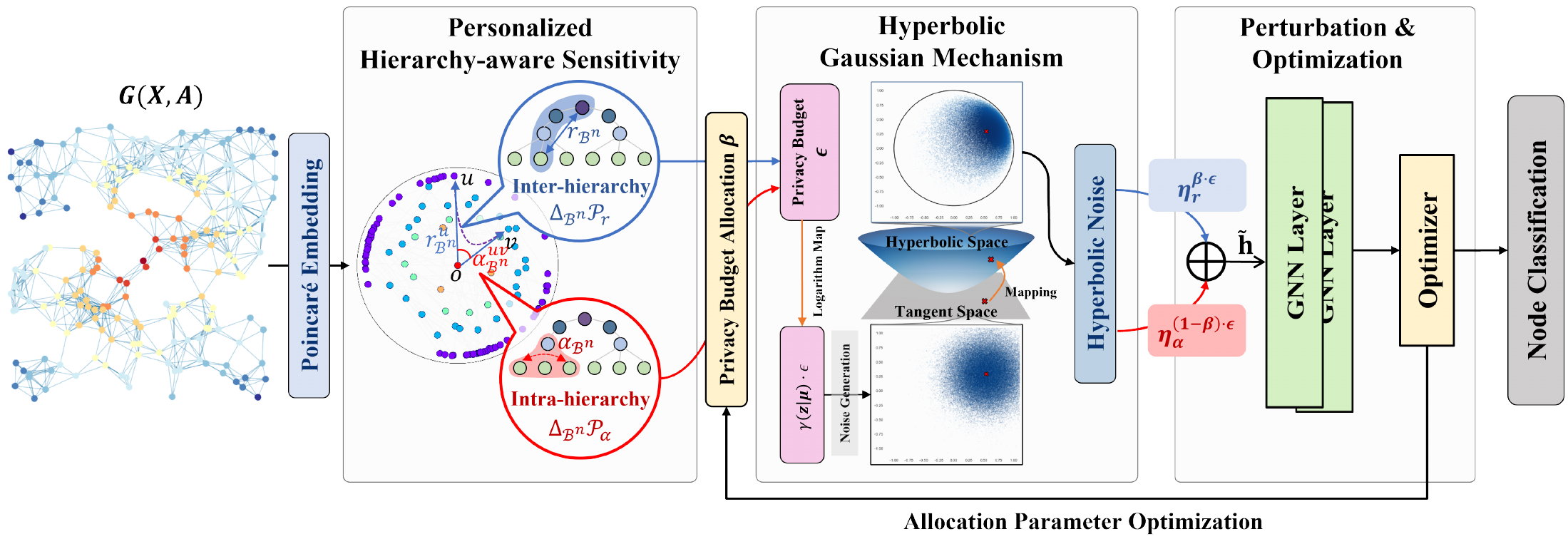}
    \caption{
    Overview of \modelname.
    Given $\mathcal{G}$ as input, \modelname~consists of the following three steps: 
    (1) PHS Computing: We first obtain the Poincar\'e embedding using the adjacency matrix $A$ and compute the PHS.
    (2) Noise generation: The sensitivity is utilized to perform the HGM in order to obtain the hyperbolic noise satisfying $(\epsilon,\delta)$-DP. 
    (3) Perturbation \& Optimization: The noise is injected into GNNs, and the privacy budget allocation is optimized according to the downstream task feedback.}
\label{fig:framework}
\end{figure*}

\section{Our Approach}
\label{sec:typestyle}
In this section, we introduce the overall learning framework of \modelname, a unified hierarchy-aware graph neural network for privacy guarantees with differential privacy, and find out how to achieve privacy protection in the hierarchy structure. 
The framework is shown in Figure~\ref{fig:framework}, and the overall process of \modelname~is shown in Algorithm~\ref{Alg:algorithm}. 

\subsection{Personalized Hierarchy-aware Sensitivity}
To the best of our knowledge, as described in many existing privacy-preserving models, the sensitivity of differential privacy is usually measured by the $L_2$ norm (Euclidean distance), which makes it difficult to measure the non-Euclidean structure accurately. 
Therefore, the sensitivity of the traditional method measured under Euclidean space is not accurate when used in the hierarchy.
Moreover, to solve the personalized privacy requirements of hierarchical structures, we aim to be able to explicitly design for inter- and intra-hierarchy sensitivity.
Inspired by the hyperbolic geometric prior, we design a novel \textit{Personalized Hierarchy-aware Sensitivity} based on the Poincar\'e embedding~\cite{NickelK17Poincare} for generating random perturbation noise with adaptive inter- and intra-hierarchy correlations.

\textbf{Hierarchy-aware node representation.}
First, we need to explicitly represent the graph hierarchy based on the Poincar\'e embedding. 
We utilize the Poincar\'e embedding to learn a hierarchy-aware node representation, which is a shallow model, and minimize a hyperbolic distance-based loss function. 
Then We learn node embeddings $\mathbf{e}^{V} =\left \{ \mathbf{e}_{i} \right \} _{i=1}^{\| \mathcal{V} \|}$ ($\mathbf{e}_{i}\in \mathcal{B}^{n}, V\in \mathcal{V} $)\footnote{We use the Poincar\'e ball model with standard constant negative curvature $\|c\| = 1$, the curvature parameter $c$ will be omitted in our method.} which represents the hierarchy of nodes in the Poincar\'e ball model based on Eq.~\eqref{Eq:hyperbolic_distance}. 
The embeddings can be optimized as 
\begin{equation}\label{eq:optimize}
\begin{aligned}
\mathbf{\Theta}^{\prime} \gets & \underset{\mathbf{\Theta}} {\mathrm{argmin}} \mathcal{L(\mathbf{\Theta})}, 
&\mathbf{s.t.}~\forall \theta_{i} \in \mathbf{\Theta}:\left\|\theta_{i}\right\| < {1/c}, 
\end{aligned}
\end{equation}
where $\mathcal{L}(\mathbf{\Theta})$ is a softmax loss function that approximates the dependency between nodes, $\mathbf{\Theta}$ is the parameters of Poincar\'e ball model.

We can obtain the radius and angle of the node on the Poincar\'e disk based on the Poincar\'e embedding $\mathbf{e}^{V}$. 
A smaller $\mathrm{Norm}_{\mathcal{B}^{n}}(\mathbf{e}_u)$ indicates that $u$ is located at the top of the hierarchy.
The node with a top-level hierarchy is approximately near the center of the disk and plays a more important role in the graph. 
Then we can give the inter-hierarchy sensitivity by using the radius $r$ of nodes on Poincar\'e disk.

\begin{myDef}[Inter-hierarchy Sensitivity]\label{def:inter-hierarchy}
Given $V$ and $V^{\prime}$ are the neighboring subsets of graph nodes, and $V$ and $V^{\prime}$ only differ by one node. The inter-hierarchy sensitivity can be defined as: 
\begin{equation}
\begin{aligned} \label{eq:r_sens}
    \Delta  _{\mathcal{B}^n}\mathcal{P}_r =\max_{V,V^{\prime}} \left | \mathrm{Norm}_{\mathcal{B}^{n}}(\mathbf{e}^V) -\mathrm{Norm}_{\mathcal{B}^{n}}(\mathbf{e}^{V^{\prime}}) \right |.
\end{aligned}
\end{equation}
\end{myDef}

On the other hand, since the angle sector on the hyperbolic disk indicates the node similarity or the community, we use it to measure the correlations of nodes within a hierarchy level.
As the Poincar\'e ball is conformal to Euclidean space~\cite{HNN:GaneaBH18}, the angle between two vector $u,v$ at the radius $r$ is given by
\begin{equation}
\begin{aligned} \label{eq:angle}
   \alpha(u,v)|_{\mathbf{e}^{V}}=\frac{g_{x}^{\mathcal{B} ^{n}}(u,v)}{\sqrt{g_{x}^{\mathcal{B} ^{p}}(u,u)}\sqrt{g_{x}^{\mathcal{B} ^{n}}(v,v)}} =\frac{\left \langle \mathbf{e}_u,\mathbf{e}_v \right \rangle }{\left \| \mathbf{e}_ u \right \|\left \| \mathbf{e}_v \right \|  }  .
\end{aligned}
\end{equation}
Similarly, we measure the correlation within the intra-hierarchy based on the angle $\alpha$ between any two nodes on the Poincar\'e disk.
\begin{myDef}[Intra-hierarchy Sensitivity]\label{def:intra-hierarchy}
Given $V$ and $V^{\prime}$ are the neighboring subsets of graph nodes, and $V$ and $V^{\prime}$ only differ by one node. The intra-hierarchy sensitivity can be defined as: 
\begin{equation}
\begin{aligned} \label{eq:a_sens}
    \Delta  _{\mathcal{B}^n}\mathcal{P}_\alpha =\max_{V,V^{\prime } }   \left \| \alpha \left (V,V^{\prime } \right ) |_{\mathbf{e}^{(V \cup V^{\prime})} }  \right \|_{\mathcal{B}^{n}}.
\end{aligned}
\end{equation}
\end{myDef}

Then we utilize the inter-hierarchy $\Delta _{\mathcal{B}^n}\mathcal{P}_r$ and intra-hierarchy $\Delta _{\mathcal{B}^n}\mathcal{P}_\alpha$ sensitivities separately to focus on the importance of nodes at different radius and angles and generate perturbation noises that satisfy personalization.

\subsection{Hyperbolic Gaussian Mechanism}\label{subsec:HGM}
The existing works widely used differential privacy strategies based on Laplace noise or Gaussian noise to achieve protection.
However, due to the difference in metric scales, their noise computation can only be performed in flat Euclidean space, which is difficult to adapt to curved hyperbolic space.
To address the privacy issues proposed by the hierarchy of graphs, we design a \textit{Hyperbolic Gaussian Mechanism} that will extend the Gaussian mechanism in Euclidean space to hyperbolic space based on the \textit{Wrapped Gaussian Distribution}~\cite{wrapped-gaussian} to realize stochastic perturbations that satisfy differential privacy in the metric of hyperbolic space. 
The hyperbolic Gaussian distribution with $c=1$ is defined as 
\begin{equation}\label{eq:curv-noise}
\begin{aligned}
&\mathcal{N} _{\mathcal{B}^n}(\mathbf{z}|\mathbf{\mu} ,\sigma_{\epsilon }^2 \mathbf{I}   )=\mathcal{N}(\lambda _{\mathbf{\mu }} \log_{\mathbf{\mu}}(\mathbf{z})|\mathbf{0},\sigma_{\epsilon }^2 \mathbf{I} ) \cdot \gamma((\mathbf{z}|\mathbf{\mu})),
\\
&\text{with}~\gamma(\mathbf{z}|\mathbf{\mu}) =\left ( \frac{d_{\mathcal{B}^n} (\mathbf{\mu} ,\mathbf{z} ) }{\sinh d_{\mathcal{B}^n (\mathbf{\mu} ,\mathbf{z} )}}  \right )^{n-1},
\end{aligned}
\end{equation}
where $\mu \in \mathcal{B} ^n$ is mean parameter, $\sigma_{\epsilon  }\in \mathbf{R}^n$ is standard deviation, $\log_{\mu}(\cdot)$ is the logarithm map function,and $\gamma$ represents the spatial mapping and normalization.

\noindent\textbf{Hyperbolic Gaussian Mechanism.}
Let $f:\mathcal{B}^{|\mathcal{X}|} \rightarrow \mathbf{R}^n$ be an arbitrary $n$-dimensional function, and define its hyperbolic sensitivity to be $\Delta_{\mathcal{B}^n}f = \max_{\mathrm{adjacent}(D,D^{\prime})} \| f(D) - f(D^{\prime}) \|_{\mathcal{B}^n}$. 
The \textit{Hyperbolic Gaussian Mechanism} with parameters $\sigma$ adds noise scaled to $\mathcal{N} _{\mathcal{B}^n}(\cdot|0 ,\sigma^2 \mathbf{I})$ to each of the $n$ components of the output.
\begin{thm}\label{thm:1}
Let $\epsilon \in (0,1)$ be arbitrary. For $c^2 > 2 \ln(1.25 \gamma(\cdot|\mu) /\delta)$, the Hyperbolic Gaussian Mechanism with parameter $\sigma \ge c \log_{\mu} (\Delta_{\mathcal{B}^n}f)  \gamma(\cdot|\mu)/  \epsilon $ is $(\epsilon,\delta)$-differentially private on hyperbolic space.
\end{thm}
To satisfy the $(\epsilon,\delta)$-differentially private in hyperbolic space, the hyperbolic sensitivity and hyperbolic Gaussian noise sampling need to be mapped to the tangent space by logarithm map function $\log_{\mu}(\cdot)$, and the privacy budget $\epsilon$ and parameter $\delta$ also need to be isometric mapped in the tangent space of $\mu$. 
Please refer to Appendix A.1 
for the detailed proof. 

According to the above, we can obtain two kinds of perturbation noise based on the inter-hierarchy $\Delta _{\mathcal{B}^n}\mathcal{P}_r$ and intra-hierarchy $\Delta _{\mathcal{B}^n}\mathcal{P}_\alpha$ sensitivities. The Hyperbolic Gaussian Noise can be generated by
\begin{equation}\label{eq:noise}
\begin{aligned}
\eta_{r}^{\epsilon_{r}} &\sim \mathcal{N} _{\mathcal{B}^n}(\mathbf{z}|\mathbf{\mu} ,c^2 \log_{\mu} (\Delta_{\mathcal{B}^n}\mathcal{P}_r)^2  \gamma(\mathbf{z}|\mu)^2/\epsilon_{r}^2 \mathbf{I}), \\
\eta_{\alpha}^{\epsilon_{\alpha}} &\sim \mathcal{N} _{\mathcal{B}^n}(\mathbf{z}|\mathbf{\mu} ,c^2 \log_{\mu} (\Delta_{\mathcal{B}^n}\mathcal{P}_\alpha)^2  \gamma(\mathbf{z}|\mu)^2/\epsilon_{\alpha}^2 \mathbf{I}). \\
\end{aligned}
\end{equation}

\subsection{Perturbation and Optimization}
To better utilize hierarchical information to provide hierarchy-aware privacy perturbations, we utilize GNNs to capture the domain representation of nodes. Given a graph $\mathcal{G}=\left ( \mathcal{V},\mathcal{E} \right )$ with node set $\mathcal{V}$ and edge set $\mathcal{E}$. 
For the semi-supervised node classification task, given the labeled node set $\mathcal{V}_L$ and their labels $\mathcal{Y}_L$, where each node $v_i$ is mapped to a label $y_i$, our goal aims to train a node classifier $f_{\theta}$ to predict the labels $\mathcal{Y}_U$ of remaining unlabeled nodes $\mathcal{V}_U=\mathcal{V} \setminus \mathcal{V}_L$. 
Therefore, following the aggregation and update mechanism of message passing in GNNs, we define the embedding learning of nodes $u$ in $(l+1)$-th layer as
\begin{equation}\label{eq:covn}
\begin{aligned}
\mathbf{h} _{u}^{(l+1)} =\sigma \left(\sum_{v\in \mathcal{V}(u) }c_{v} \mathbf{W}^{(l)} \mathbf{h}_{v}^{(l)} \right),
\end{aligned}
\end{equation}
where $c_v$ is a node-wise normalization constant and $\mathcal{V}(u)$ is the neighbor set. 
During the continuous iteration in the training stage, the features of node $u$ will be updated with a hyperbolic Gaussian mechanism as 
\begin{equation}
\begin{aligned} \label{eq:perturb}
&\hat{\mathbf{h}} = \mathbf{h}+  \eta_{r}^{\beta \cdot \epsilon} +  \eta_\alpha^{{(1-\beta) \cdot \epsilon}},
\end{aligned}
\end{equation}
where $\beta$ is the normalized attention weight to learn the inter- and intra-hierarchy importance in nodes and rationally allocate the privacy budget, i.e. $\epsilon_r +\epsilon_\alpha =\epsilon$. 
 
Therefore, we complete the noise generation and addition by hierarchy-aware mechanism. 
The objective for \modelname~is the average loss of predicting labels of unlabeled nodes, formulated as
\begin{equation}
\begin{aligned} \label{eq:loss}
\mathcal{L} =\frac{1}{\left \| \mathcal{V}_U  \right \| }\sum_{v\in \mathcal{V}_U }\mathcal{L}_\mathcal{G} (\mathbf{\hat{h}} _{u,v}, y_{v} ), 
\end{aligned}
\end{equation}
where $\mathcal{L}_{\mathcal{G}}$ stands for the loss of semi-supervised node classification and is implemented by cross-entropy in this work.

\begin{algorithm}[t]
    \LinesNumbered
    \caption{Overall training process of~\modelname} 
    \label{Alg:algorithm}
    \KwIn{
    Graph $\mathcal{G}=\{\mathcal{V},\mathcal{E}\}$ with node labels $\mathcal{Y}$; 
    Number of training epochs $E$.}
    \KwOut{Predicted label $\hat{\mathcal{Y}}$.}
    Parameter $\Theta $ initialization;\\
    Learning and optimizing node Poincar\'e embedding $\mathbf{e}^V \gets$ Eq.~\eqref{eq:optimize} and ~\eqref{eq:angle};\\
    \For{$e=1,2,\cdots,E$}{
        \tcp{Personalized Hierarchy-aware Sensitivity}
        Calculate hierarchy-aware sensitivity $\Delta  _{\mathcal{B}^n}\mathcal{P}_r $ and $\Delta  _{\mathcal{B}^n}\mathcal{P}_\alpha \gets$ Eq.~\eqref{eq:r_sens} and ~\eqref{eq:a_sens}; \\
        \tcp{Hyperbolic Gaussian Mechanism}
        Calculate the hyperbolic Gaussian distribution $\mathcal{N} _{\mathcal{B}^n}(\mathbf{z}|\mu ,\sigma_{\epsilon }^2 \mathbf{I}   )\gets$ Eq.~\eqref{eq:curv-noise};\\
        Learning node embeddings $\mathbf{h} _u \gets$ Eq.~\eqref{eq:covn};\\
        Perturbing node embeddings $\mathbf{\widehat{h} } $ by hyperbolic Gaussian noise $\gets$ Eq.~\eqref{eq:perturb}; \\
        Predict node labels $\hat{\mathcal{Y}}$ and calculate the classification loss $\mathcal{L} \gets $ Eq.~\eqref{eq:loss};\\
        Update model parameters by minimizing $\mathcal{L}$. 
    }  
\end{algorithm}
\section{Experiments}
\begin{table*}[ht]


\resizebox{\linewidth}{!}{%
\centering
\begin{tabular}{ccccccccccc}
\toprule
\multirow{2}{*}{\raisebox{-1ex}{Model}}
& \multicolumn{2}{c}{Cora} 
& \multicolumn{2}{c}{Citeseer} 
& \multicolumn{2}{c}{PubMed}
& \multicolumn{2}{c}{Computers}
& \multicolumn{2}{c}{Photo}
\\ \cmidrule{2-11}

&  W-F1 &  M-F1 &  W-F1 &  M-F1 &  W-F1 &  M-F1 &  W-F1 &  M-F1 &  W-F1 &  M-F1
\\ \midrule
GCN    &80.0±1.1 &80.1±1.1 &68.1±0.2 &68.6±0.2 &\underline{78.5±0.5}  &\underline{78.5±0.5}  &84.7±2.3  &82.5±3.6  &90.2±1.4  &89.6±1.6 \\
GAT    &\underline{81.6±1.1} &\underline{81.8±1.0} &\underline{69.4±1.2} &\underline{70.0±1.0}  &77.0±0.5  &77.0±0.4  &\underline{87.5±0.4}  &\underline{87.1±0.5}  &\textbf{92.9±0.2}  &\textbf{92.8±0.2}  \\
HyperIMBA    &\textbf{83.0±0.3}  &\textbf{83.1±0.4}  &\textbf{76.3±0.2}  &\textbf{73.4±0.3}  &\textbf{86.6±0.1}  &\textbf{86.5±0.1}  &\textbf{89.6±0.2}  &\textbf{89.6±0.1}  &\underline{92.8±0.3}  &\underline{92.5±0.3} \\
\specialrule{0em}{1.5pt}{1.5pt}
\midrule
\midrule
\specialrule{0em}{1.5pt}{1.5pt}
VANPD    &40.9±1.6  &41.5±1.6  &35.6±1.2  &35.6±1.2   &61.8±0.2  &61.8±0.3  &74.1±1.1  &74.3±1.0  &84.4±1.0  &84.3±1.1  \\
LaP     &62.6±0.9  &61.4±0.9  &55.0±1.5  &53.2±1.5   &68.3±0.2  &68.2±0.2  &80.1±1.0  &\underline{79.9±1.0}  &88.9±0.9  &88.7±1.0\\
RdDP    &78.1±0.2 &75.1±0.4 &73.1±0.5 &70.0±0.7  &79.1±0.7  &78.6±0.9  &80.5±0.9  &76.1±1.6  &91.4±0.2  &90.1±0.5 \\
AtDP    &\textbf{81.0±0.2} &\textbf{80.0±0.2} &\underline{74.8±0.1} &\underline{72.0±0.2} &\underline{83.5±0.0}  &\underline{83.5±0.0}  &\underline{81.5±4.4} &78.4±7.2  &\underline{91.7±0.6}  &\underline{91.3±0.7}\\
\textbf{\modelname}    &\underline{78.2±0.6} &\underline{75.5±1.2} &\textbf{75.5±0.2} &\textbf{72.5±0.2}  &\textbf{83.8±0.2}  &\textbf{83.7±0.2}  &\textbf{86.9±0.4}  &\textbf{86.5±0.5}  &\textbf{92.6±0.2}  &\textbf{92.4±0.3}   \\
\bottomrule
\end{tabular}
}
\caption{Weighted-F1 and Micro-F1 score of the node classification task. (Result: average score ± standard deviation; Bold: the best of baseline model; Underline: runner-up.)}
\label{tab:summaryResult}
\end{table*}

\begin{table*}[ht]

\resizebox{\linewidth}{!}{%
\centering
\begin{tabular}{ccccccccccc}
\toprule
\multirow{2}{*}{\raisebox{-1ex}{Model}}
& \multicolumn{2}{c}{Cora} 
& \multicolumn{2}{c}{Citeseer} 
& \multicolumn{2}{c}{PubMed}
& \multicolumn{2}{c}{Computers}
& \multicolumn{2}{c}{Photo}
\\ \cmidrule{2-11}

& W-F1   & $\Delta$ (\%)  
& W-F1   & $\Delta$ (\%)  
& W-F1   & $\Delta$ (\%)  
& W-F1   & $\Delta$ (\%)  
& W-F1   & $\Delta$ (\%)  
\\ \midrule
\textbf{\modelname~}   &\textbf{59.9±1.4}  &-  &\textbf{74.0±1.3}  &-  &\textbf{79.4±0.5}  &-  &\textbf{83.8±0.4}  &-  &\textbf{91.9±0.5}  &-  \\
\modelname~(\textit{w/o inter})    &48.4±2.6  &$\downarrow$11.5 &60.6±4.1  &$\downarrow$13.4  &70.8±1.4  &$\downarrow$8.6  &79.5±1.8  &$\downarrow$4.3  &91.3±0.4  &$\downarrow$0.6 \\
\modelname~(\textit{w/o intra}) &48.8±0.4  &$\downarrow$11.1  &60.1±9.9  &$\downarrow$13.9  &76.6±1.4  &$\downarrow$2.8  &82.6±1.2  &$\downarrow$1.2  &91.3±0.2  &$\downarrow$0.6  \\
\modelname~(\textit{w/o allocate}) &51.2±2.4  &$\downarrow$8.7  &69.5±2.8  &$\downarrow$4.5  &77.2±0.7  &$\downarrow$2.2  &82.7±0.4  &$\downarrow$1.1  &91.5±0.4  &$\downarrow$0.4   \\
\bottomrule
\end{tabular}
}
\caption{Weighted-F1 scores (\% ± standard deviation) and improvements (\%) results of Ablation Study. (Result: average score ± standard deviation; Bold: best.) }
\label{tab:ablation}
\end{table*}

In this section, we conduct experiments on five datasets and seven baselines to demonstrate the privacy protection adaptability and the graph learning effectiveness of \modelname~based on a semi-supervised node classification task. 
\subsection{Dataset and Model Setup}
\textbf{Datasets.} 
For datasets (see Appendix B.1), we chose three citation networks (Cora, Citeseer and PubMed) and two E-commerce networks in Amazon (Computers and Photo). \\

\noindent\textbf{Baselines.} For baselines, GCN~\cite{GCN}, GAT~\cite{GAT}, and HyperIMBA~\cite{HyperIMBA} are convolutional neural networks model, attention neural networks model, and hierarchy-aware model, respectively. 
VANPD and LaP~\cite{MIA3_TPS-ISA2021} which use the Laplace noise perturbation mechanism are privacy models in Euclidean space.
RdDP, AtDP, and the proposed~\modelname~are privacy methods in hyperbolic spaces. 
RdDP and AtDP are two variant models of DP noise generation, representing the addition of random noise and attention-aware noise, respectively. 

\noindent\textbf{Settings. } \modelname~performs the semi-supervised node classification task to verify its privacy performance. 
Our dataset split follows the PyTorch Geometric.
The learning rate $lr$ is $0.005$, the privacy budget $\epsilon$ to be $[0, 1]$, and the training iterations $E$ to be $200$. 
For other model settings, we adopt the default optimal values in the corresponding papers. 
We conducted the experiments with NVIDIA GeForce RTX 3090 with 16GB of Memory. 

\subsection{Performance Evaluation}

\noindent\textbf{Performance of Node Classification. }
We evaluate~\modelname~for node classification where privacy models are trained in $\epsilon=1$. 
The Weighted-F1 and Micro-F1 scores are reported in Table~\ref{tab:summaryResult} where the best results are shown in bold and the runner-up results are shown in underline. 
It can be observed from the results that differential privacy-based models perform worse on the classification task compared with non-DP models while increasing the protection for sensitive information, which is caused by adding extra noise.  
Notably, \modelname~gets the absolute upper hand in terms of performance among privacy-preserving models compared to other privacy-preserving models. 
Because the hyperbolic noise is more adapted to the operations in the hierarchical structure, the destructive power in the Euclidean noise is significantly attenuated, resulting in uniformly higher performance. 
In conclusion, on the premise of improving the ability for privacy protection, \modelname~preserves the data availability as much as possible and improves the performance of the node classification task.

\noindent\textbf{Ablation Study.} 
In this subsection, we conduct the ablation study for~\modelname~to validate the model utility provided by our consideration of node hierarchies (\textit{w/o inter}) and correlations (\textit{w/o intra}) on a hierarchical structure, and to remove the adaptive privacy budget allocation (\textit{w/o allocate}) to these two properties, i.e., the optimization of hyperbolic noise is removed. 
We set $\epsilon= 0.01$ for easy observation. 
The results as shown in Table~\ref{tab:ablation}, indicate that missing any component of \modelname~leads to a degradation of the performance, where \modelname~(\textit{w/o allocate}) has the smallest impact in most of the datasets, but numerically demonstrates the effectiveness of the privacy budget allocation. 
In addition, the one-sided perturbations in both \modelname~(\textit{w/o inter}) and \modelname~(\textit{w/o intra}) experiments reflect a strong influence on the model performance, suggesting that they have individualized perturbation rules for the nodes.

\begin{figure}[t]
	\centering
	\includegraphics[width=0.45\textwidth]{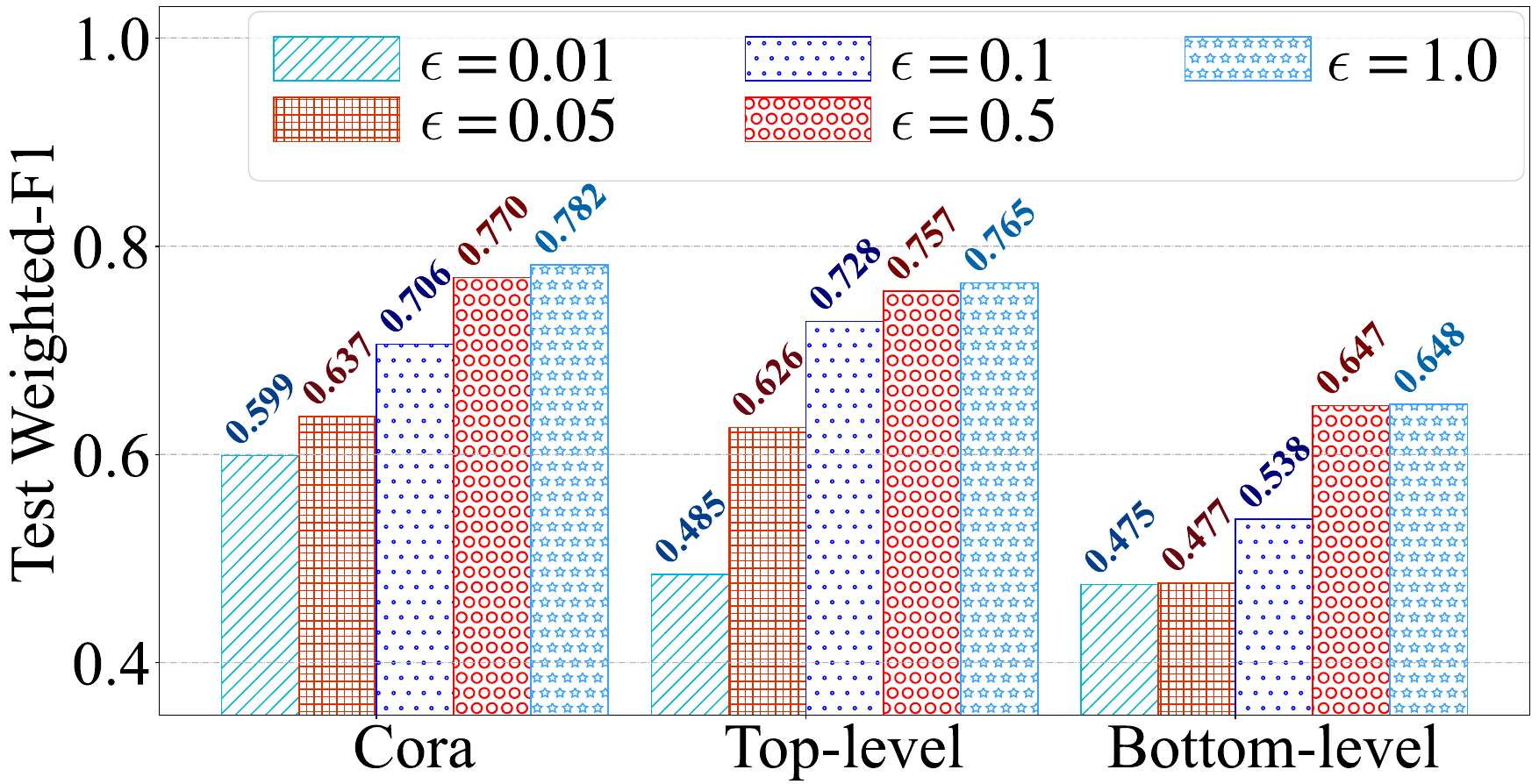}
	\caption{Hierarchical sensitivity experiments on Cora. 
}
\label{fig:sensitive}
\end{figure}

\noindent\textbf{Case Study and Analysis of Sensitivity.} 
\textit{As a case study}, to verify the effectiveness of~\modelname~in privacy protection and its generalization ability to hierarchical structures, three splits for Cora are provided as \textit{Cora} (random sampling training set), \textit{Top-level} and \textit{Bottom-level} (sampling the top 33\% and bottom 33\% of the training samples, respectively), and their training sets with moderate, weak, and strong sensitivity, respectively. 
Note that the nodes are ordered from highest to lowest according to the Poincar\'e weights, indicating that the nodes range in sensitivity from lowest to highest. 
As shown in Fig.~\ref{fig:sensitive}, Cora, which randomly samples the training nodes, has the best overall performance and is comparable to the Top-level, and finally Bottom-level. 

\textit{For the analysis of sensitivity}, we evaluate the model performance by setting different $\epsilon$ from $0.01$ to $1$, where $\epsilon$ measures the strength of the model's privacy protection, with smaller values indicating greater privacy protection power, less usability, and more information loss.  
As shown in Fig.~\ref{fig:sensitive}, for the overly strict $\epsilon$ = 0.01, both the Top-level and the Bottom-level show the worst performance that can be understood, but in the looser limits, the Top-level samples perform well (these nodes are decisive for the downstream task so the amount of perturbation is low and the performance is almost close to Cora's). 
Whereas~\modelname~in the Bottom-level samples adapts the requirement of needing a high degree of privacy preservation while providing the protection ability in a high privacy budget for sensitive data.

\begin{figure}[t]
	\centering
	\includegraphics[width=0.45\textwidth]{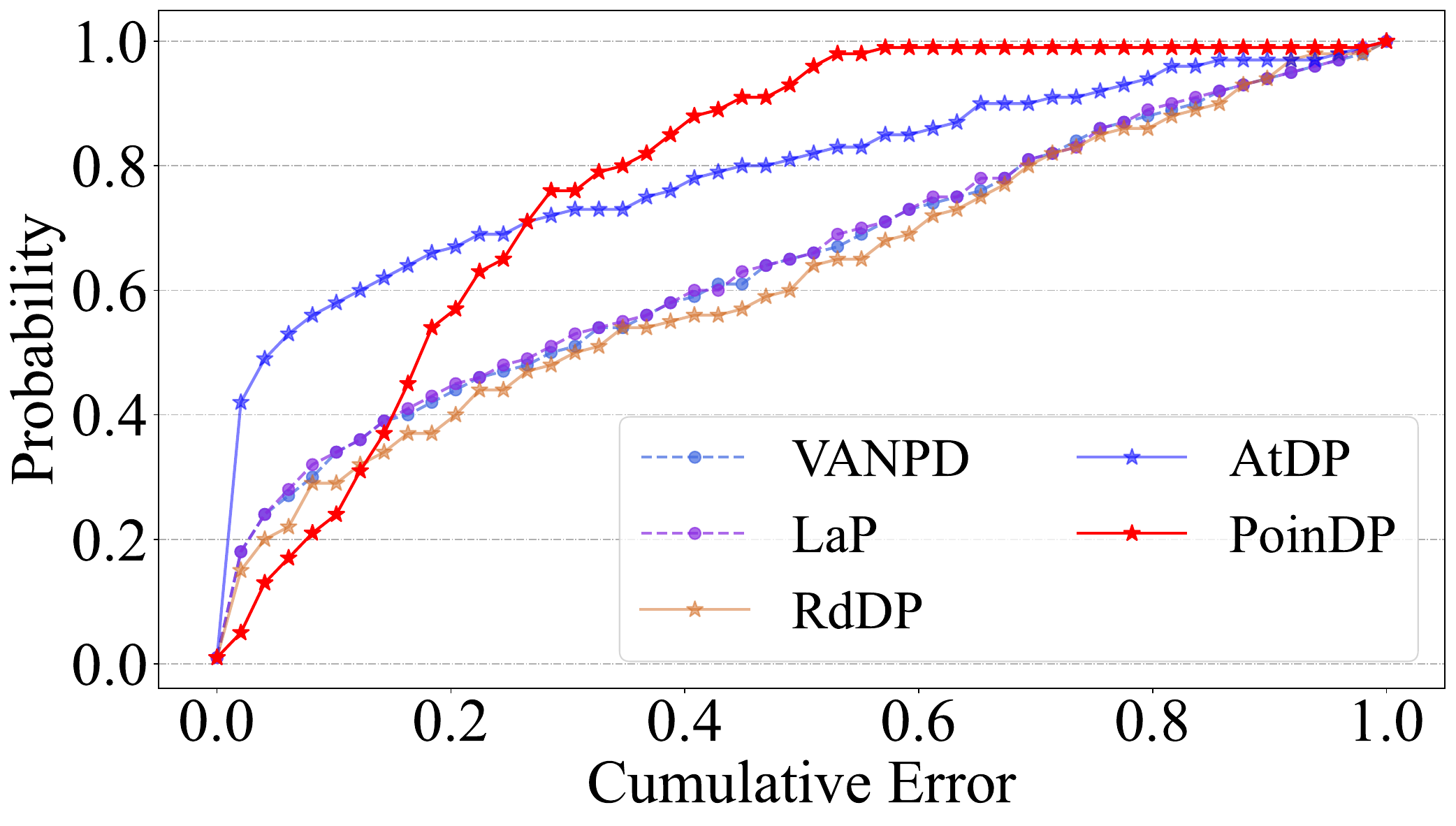}
	\caption{Cumulative error distribution with differential privacy-preserving method on Cora. 
}
\label{fig:distribution_fig}
\end{figure}

\noindent\textbf{Analysis of Noise Distribution. } Fig.~\ref{fig:distribution_fig} shows that the noise mechanism of~\modelname~by the cumulative error distribution. 
We compare five privacy-preserving models and find that the error accumulation for~\modelname~grows the fastest and ends its accumulation at 0.5, indicating a focused imposition of noise, and reflecting the individualized hierarchical perturbation mechanism of~\modelname.
However, others are slow to converge, indicating a high percentage of results with large error values, and they aimlessly put noise into the samples, leading to poor usability. 
Overall, our hyperbolic Gaussian mechanism can put noise for some samples in a focused manner, providing personalized protection capability.

\begin{figure}[t]
\centering
\subfigure[VANPD.]{
\includegraphics[width=0.45\linewidth]{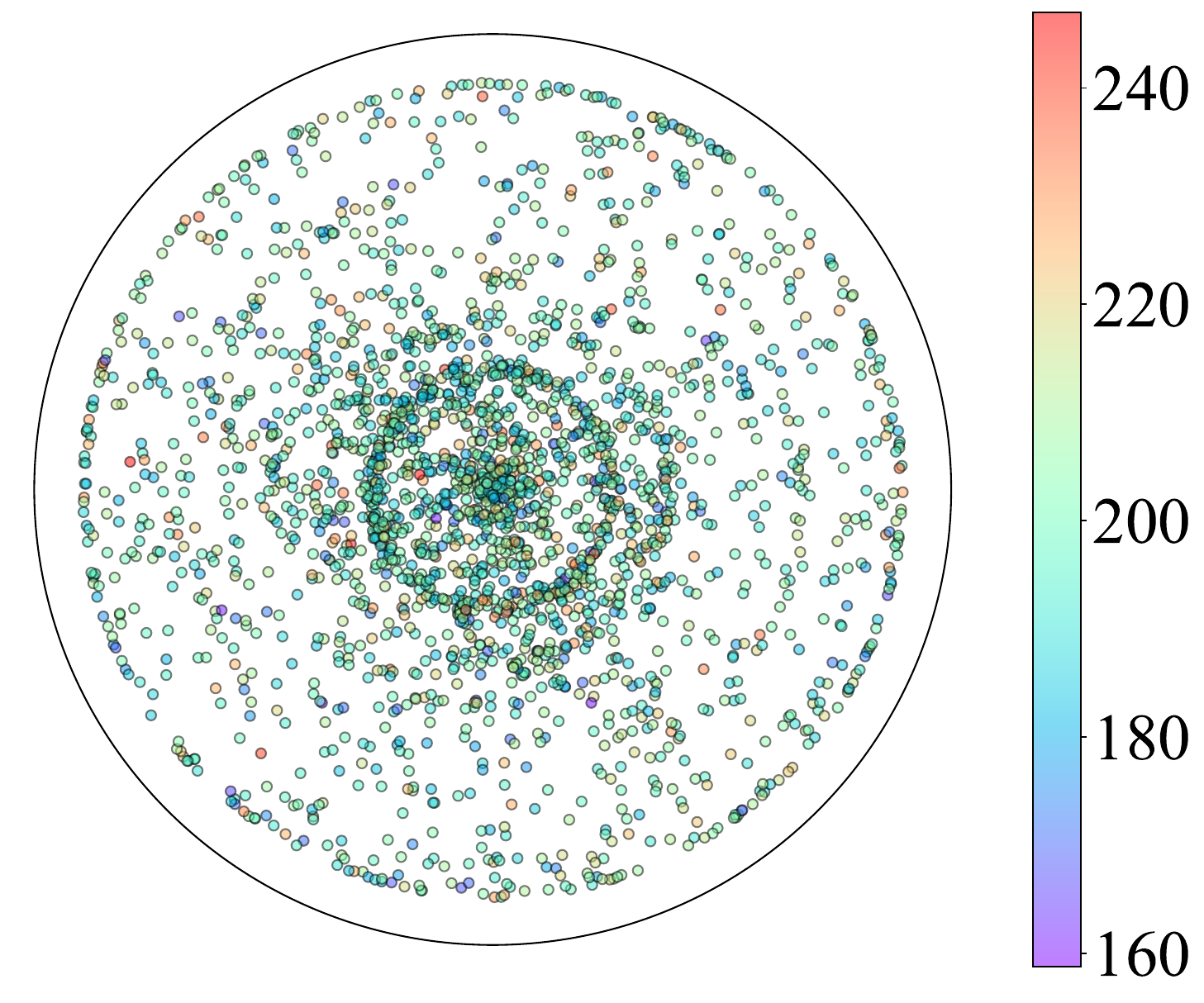}
}%
\hspace{0.2cm}
\subfigure[PoinDP.]{
\includegraphics[width=0.45\linewidth]{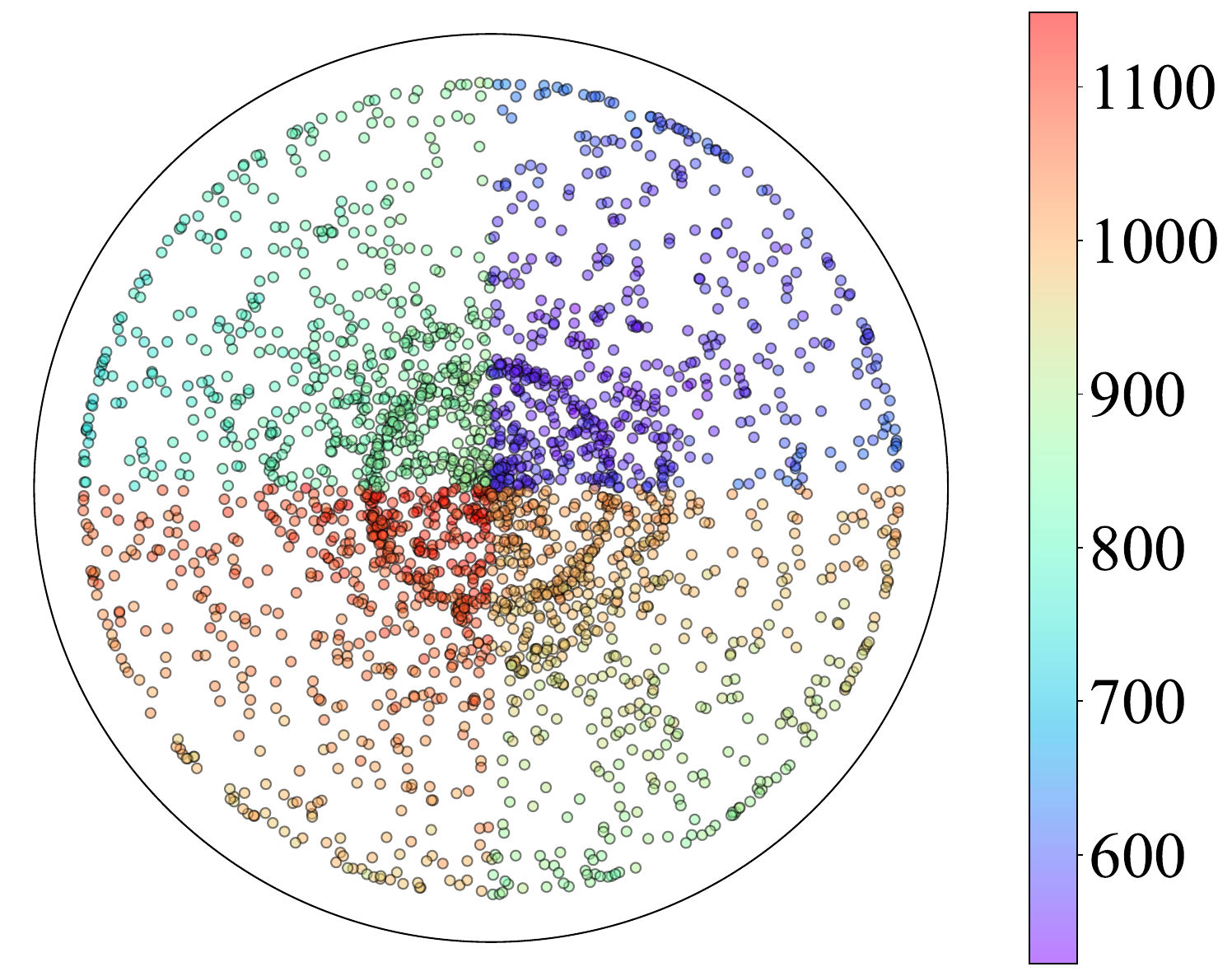}
}%
\centering
\caption{Visualization of noise distribution on Poincar\'e disk for VANPD and~\modelname~on Cora. }
\label{fig:visualization}
\end{figure}

\noindent\textbf{Visualization.} We visualize the noise distribution of the four privacy models on the Cora dataset to intuitively represent the ability of our models to perceive hierarchical structures. Please refer to Appendix B.2 
for other visualizations. 
Fig.~\ref{fig:visualization} expresses the data as its whole with a hierarchical structure, where the colors represent the amount of noise and the position of each point is the layout of the node on the poincar\'e disk. 
As can be noticed in~\modelname~in Fig.~\ref{fig:visualization} (b), as the radius of the disk increases, the noise nodes become lighter in color and exhibit different colors at different angles, which fully demonstrates~\modelname's excellent ability to capture inter- and intra-hierarchy information. 
In contrast, the other models exhibit uniform perturbations to the hierarchy.  
In a nutshell, benefiting from the PHS and HGM mechanisms, ~\modelname~again demonstrates its effectiveness. 

\textbf{Attack Experiment.} We conduct Membership Inference Attack (MIA)~\cite{MIA3_TPS-ISA2021} and the results are reported in Table~\ref{tab:attack}. 
Please refer to Appendix B.3 
for the detailed attack settings and performance analysis. 
The conclusion is that hierarchical information $\mathcal{H}$ can enhance the attacker's reasoning ability, while PoinDP can provide superior protective capabilities.

\begin{table}[t]

\resizebox{\linewidth}{!}{%
\centering
\begin{tabular}{c|c|cc|cc}
\toprule
\multicolumn{2}{c|}{\makecell{\textbf{Dataset}\\\textbf{Hyperbolicity} $\delta$ }} & \multicolumn{2}{c|}{\makecell{\textbf{PubMed}\\$ \delta = 1.65 $}} & \multicolumn{2}{c}{\makecell{\textbf{Photo}\\$ \delta = 0.15 $}}\\
\cmidrule{1-6}
\multicolumn{2}{c|}{\textbf{Metric} }
& AUC $\uparrow $ &Prec. $\uparrow $ & AUC $\uparrow $ &Prec. $\uparrow $
\\ \midrule
\multicolumn{1}{c|}{\multirow{3}[6]{*}{\raisebox{2.5em}{\rotatebox{90}{Attack   }}}} & GCN    &62.8±1.5  &63.8±1.4    &79.7±2.5  &80.7±2.6  \\
& GAT    &58.4±2.4  &59.3±2.8   &79.7±0.9 &80.4±0.8    \\
& GCN+$\mathcal{H}$   &\textbf{63.2±0.2}  &\textbf{64.0±0.2}  &\textbf{82.4±0.4}   &\textbf{83.4±0.4}   \\
\specialrule{0em}{1.5pt}{1.5pt}
\midrule
\midrule
\specialrule{0em}{1.5pt}{1.5pt}
\multicolumn{2}{c|}{\textbf{Metric} }
& AUC $\downarrow $ &Prec. $\downarrow $   & AUC $\downarrow $  &Prec. $\downarrow $  \\
\midrule
\multicolumn{1}{c|}{\multirow{3}[6]{*}{\raisebox{3em}{\rotatebox{90}{Defense}}}} & VANPD  &51.1±0.3  &51.2±0.5   &68.1±1.6   &68.9±1.5  \\
& LaP    &51.9±0.5  &52.8±0.5  &70.3±0.1   &71.0±0.1   \\
& \textbf{\modelname}    &\textbf{46.7±2.1}  &\textbf{46.5±3.0}  &\textbf{37.4±0.6}  &\textbf{34.8±2.0}   \\
\bottomrule
\end{tabular}
}
\caption{Membership Inference Attack (MIA) performance. \\
($\uparrow $: the higher, the better; $\downarrow $: the lower, the better) }
\label{tab:attack}
\end{table}

\section{Conclusion}
In this paper, for the first time, we propose the privacy leakage problem caused by the hierarchical structure of the graph and define the problem from the perspective of hyperbolic geometry. 
We propose~\modelname, a novel and unified privacy-preserving graph learning framework for the hierarchical privacy leakage issue. 
\modelname~designs personalized hierarchy-aware sensitivities and defines differential privacy techniques in hyperbolic space, and obtains high-quality graph representations while satisfying privacy guarantees.
Experimental results empirically demonstrate the superior hierarchy perception capability of our framework and obtain excellent privacy preservation.

\section{Acknowledgments}
The corresponding authors are Xingcheng Fu and Chunming Hu. This paper is supported by the National Key Research and Development Program of China Grant (No. 2022YFB4501901) and the National Natural Science Foundation of China (No. U21A20474 and 62302023). We owe sincere thanks to all authors for their valuable efforts and contributions. 
\bibliography{main}
\appendix
\clearpage
\setcounter{secnumdepth}{2}
\setcounter{thm}{0}  
\section{Proofs}
\subsection{Proof of Theorem~\ref{thm:1}}
\label{prof:thm1}
The proof of Theorem~\ref{thm:1} in Section~\ref{subsec:HGM}.
\begin{proof}
Given a database $\mathcal{D}$ and a query $f$, the perturbation mechanism will return $f(\mathcal{D})+\eta$, where the noise is normally distributed in hyperbolic space. We are adding noise $\mathcal{N} _{\mathcal{B}^n}(\cdot|0 ,\sigma^2 \mathbf{I})$ and the sensitivity for two adjacent dataset $x$ and $y$ is defined as:
\begin{equation}
\begin{aligned}
    \Delta_{\mathcal{B}^n}f=\max _{\mathrm {adjacent } (x, y)}\|f(x)-f(y)\|_{\mu}. 
\end{aligned}
\end{equation}
According to Eq.~\eqref{def:dp}, we consider the conditions $\sigma$ under which the privacy loss is bounded by $\epsilon$. 
For convenience, we denote $\|\cdot\|_{\mu}$ as the \textbf{Poincar\'e Norm} on the pole $\mu$. 
Let $v$ be any vector satisfying $\|v\|_{\mu} \le \Delta_{\mathcal{B}^n}f$. 
For a fixed pair of databases $x, y$, we have $v = f(x) - f(y)$.
The privacy loss in hyperbolic space is
\begin{equation}
\begin{aligned}
    \left|\ln \frac{\mathrm{e}^{\left(-1 / 2 \sigma^{2}\right)\|x\|_{\mu}^{2}}}{\mathrm{e}^{\left(-1 / 2 \sigma^{2}\right)\|x \oplus v\|_{\mu}^{2}}}\right|,
\end{aligned}
\end{equation}
where the $x$ is sampled from $\mathcal{N} _{\mathcal{B}^n}(\cdot|0 ,\sigma_{\epsilon }^2 \mathbf{I})$, and the $\mu = (0,...,0)$ is the \textit{pole} of hyperbolic and tangent space.
Then we have
\begin{equation}
\begin{aligned}
\left|\ln \frac{\mathrm{e}^{\left(-1 / 2 \sigma^{2}\right)\|x\|_{\mu}^{2}}}{\mathrm{e}^{\left(-1 / 2 \sigma^{2}\right)\|x \oplus v\|_{\mu}^{2}}}\right|  =\left | \frac{1}{2 \sigma^{2}}\left(\|x\|_{\mu}^{2}-\|x \oplus v\|_{\mu}^{2}\right)\right| .
\end{aligned}
\end{equation}

We take the one-dimensional noise with an orthogonal basis as a case for high-dimensional vectors.
For the vector in hyperbolic space, a basis $b_1, ... ,b_m$, the defining $x_i=\lambda_i b_i, \lambda_i \sim \mathcal{N} _{\mathcal{B}^n}(\cdot|0 ,\sigma^2 \mathbf{I})$, and let $x=-x_0+\sum_{i=1}^{n-1} x_i$. 
According to differential geometry, we can map the vector from hyperbolic space into the tangent space of the pole $\mu$ by using the exponential map and the logarithm map, and we have
\begin{equation}
\begin{aligned}
&\left | \frac{1}{2 \sigma^{2}}\left(\|x\|_{\mu}^{2}-\|x \oplus v\|_{\mu}^{2}\right)\right| \\
= &\left |\exp_{\mu} \left( \frac{1}{2 \sigma^{2}}\left( \log_{\mu}( \|x\|^{2}- \|x + v\|^{2}\right)\right)\right|.
\end{aligned}
\end{equation}
Thus we can consider $x+v$ in the tangent space. 
Assume $v$ is orthogonal to some basis $b_j$, and we have $\|x+v\|^2 = \|v+x_j\|^2+\sum_{\{i|i\in [n] , i \ne j \} } \|x_i\|^2$ .
Thus, $\log_{\mu} \|x\|^{2}- \log_{\mu} \|x \oplus v\|^{2}) = \log_{\mu}\|v\|^2+2\cdot\lambda\cdot \log_{\mu}\|v\|$, where the parallel transformation of $x$ in the pole $\mu$ is equivalent. 
Recall that $\|v\|_{\mu} \le \Delta_{\mathcal{B}^n}f$, we can obtain
\begin{equation}
\begin{aligned}
 &\left | \frac{1}{2 \sigma^{2}}\left( \log_{\mu} \|x\|_\mu^{2}- \log_\mu \|x \oplus v\|_\mu^{2}\right)\right| \\
 \le &\left|\frac{1}{2 \sigma^{2}}\left(2  \gamma_{\mu} \lambda_{j} \log_{\mu} \Delta_{\mathcal{B}^n} f-(\log_{\mu}\Delta_{\mathcal{B}^n} f)^{2}\right)\right|.
\end{aligned}
\end{equation}
in tangent space of pole $\mu$.

For the privacy budget $\epsilon$, according to the Eq.~\eqref{eq:curv-noise}, we can isometric map $\epsilon$ to $\gamma_{\mu}^{-1} \cdot \epsilon$.
This quantity is bounded by $\epsilon$ whenever $x < \sigma^2 \epsilon / \gamma_{\mu} \log_{\mu}\Delta_{\mathcal{B}^n} f - \log_{\mu}\Delta_{\mathcal{B}^n} f /2$.
To ensure privacy loss bounded by $\epsilon$  with probability at least $1-\delta/\gamma_{\mu}$ in tangent space with the parameter $\delta$ in hyperbolic space, we require
\begin{equation}
\begin{aligned}
\mathbf{Pr}[x \ge \sigma^2 \epsilon / \gamma_{\mu} \log_{\mu}\Delta_{\mathcal{B}^n} f - \log_{\mu}\Delta_{\mathcal{B}^n} f /2] \le \delta/2\gamma_{\mu} .
\end{aligned}
\end{equation}
Denote $\sigma = c \gamma_{\mu} \log_{\mu} \Delta_{\mathcal{B}^n} f / \epsilon$.
According to the tail bound of Gaussian distribution in tangent space.
Let $t = \sigma^2 \epsilon/ \gamma_{\mu} \log_{\mu} \Delta_{\mathcal{B}^n} f - \log_{\mu} \Delta_{\mathcal{B}^n} f/2$, we have
\begin{equation}
\begin{aligned}
&\ln (t / \sigma)+t^{2} / 2 \sigma^{2}>\ln (2 / \sqrt{2 \pi} \delta/ \gamma_{\mu})\\
\Leftrightarrow & \mathbf{Pr}[x >t ] \le \frac{\sigma}{\sqrt{2 \pi}} e^{-t^{2} / 2 \sigma^{2}} < \delta/\gamma_{\mu} .
\end{aligned}
\end{equation}
Since $\epsilon \le 1$ , $c \ge 1$ and $\gamma_\mu \ge 1$, we have $c-\epsilon/2 \gamma_\mu c \ge c- 1/2\gamma_\mu$, and $\ln ( t/\sigma ) = c-\epsilon/2\gamma_\mu c > 0$ provided $c \ge 3/2$.
Then we consider the $t^2/\sigma^2$ term,
\begin{equation}
\begin{aligned}
&\left(\frac{1}{2 \sigma^{2}} \frac{\sigma^{2} \epsilon}{\gamma_{\mu} \log_{\mu} \Delta_{\mathcal{B}^n} f}-\frac{\log_{\mu} \Delta_{\mathcal{B}^n} f}{2}\right)^{2} \\
 =&\frac{1}{2 \sigma^{2}}\left[\log_{\mu} \Delta_{\mathcal{B}^n} f\left(\frac{\gamma_{\mu} c^{2}}{\epsilon}-\frac{1}{2}\right)\right]^{2} \\
 =& \frac{1}{2}\left[(\Delta f)^{2}\left(\frac{\gamma_{\mu} c^{2}}{\epsilon}-\frac{1}{2}\right)\right]^{2}\left[\frac{\epsilon^{2}}{\gamma_{\mu}^{2}c^{2}(\Delta f)^{2}}\right] \\
 =&\frac{1}{2}\left(\frac{\gamma_{\mu} c^{2}}{\epsilon}-\frac{1}{2}\right)^{2} \frac{\epsilon^{2}}{\gamma_{\mu}^{2} c^{2}} \\
 =&\frac{1}{2}\left(c^{2}-\epsilon/\gamma_{\mu}+\epsilon^{2} / 4 \gamma_{\mu}^{2} c^{2}\right).
\end{aligned}
\end{equation}

Since $\epsilon \le 1$ and $c\ge 3/2$, The boundary of $c$ has $c^2-\epsilon/\gamma_\mu +\epsilon^2/4\gamma_\mu^2c^2 \ge c^2 - 8/9$. 
Then, we have
\begin{equation}
\begin{aligned}
&c^{2}>2 \ln (\sqrt{2 / \pi})+2 \ln (1 / \delta)+\ln \left(e^{8 / 9}\right)\\
=&\ln (2 / \pi)+\ln \left(e^{8 / 9}\right)+2 \ln (\gamma_{\mu} / \delta),
\end{aligned}
\end{equation}
which, since$(2/\pi)e^(8/9)<1.55$, is satisfied whenever $\sigma \ge c \gamma_{\mu} \log_{\mu} \Delta_{\mathcal{B}^n} f / \epsilon$ and $c^2>2\ln(1.25\gamma_\mu/\delta)$.

\end{proof}

\clearpage
\section{Experiment} 

\subsection{Dataset Description} \label{sec:dataset_descrip}
We compiled detailed information about the dataset used in this paper in Table~\ref{tab:dataset_description}.
\begin{table}[t]
\small
\caption{Statistics of datasets.}
\vspace{-1em}
\centering

\begin{tabular}{cl|cccc}
\toprule
\multicolumn{2}{c|}{\textbf{Dataset}}  & \textbf{\#Node} & \textbf{\#Edge} & \textbf{\#Label} &\textbf{\#Avg. Deg} \\ 
\midrule
&\textbf{Cora}      & 2,708 & 5,429 & 7 & 4.01    \\
&\textbf{Citeseer}  & 3,327 & 4,732 & 6  & 2.85     \\
&\textbf{PubMed}     & 19,717  & 44,324 & 3   & 4.50   \\
&\textbf{Computers} & 13,752  & 245,861 & 10  & 35.76   \\
&\textbf{Photo}     & 7,487 & 119,043 & 8  & 31.80   \\
\bottomrule
\end{tabular}

\label{tab:dataset_description}
\end{table}

\subsection{Visualization} \label{sec:visualization}
\begin{figure}[t]
\centering
\hspace{0.2cm}
\subfigure[LaP.]{
\includegraphics[width=0.44\linewidth]{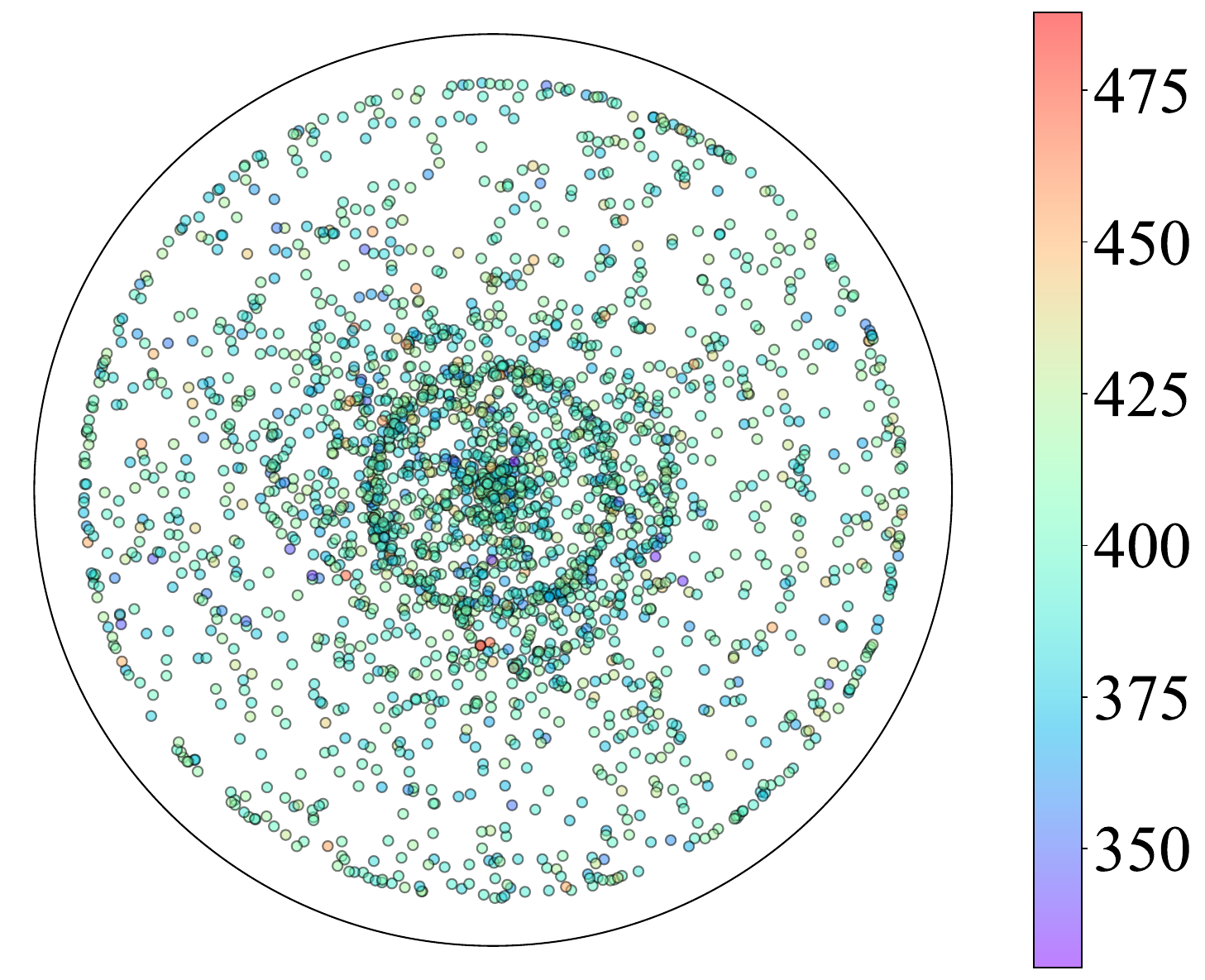}
}%
\hspace{0.2cm}
\subfigure[RdDP.]{
\includegraphics[width=0.44\linewidth]{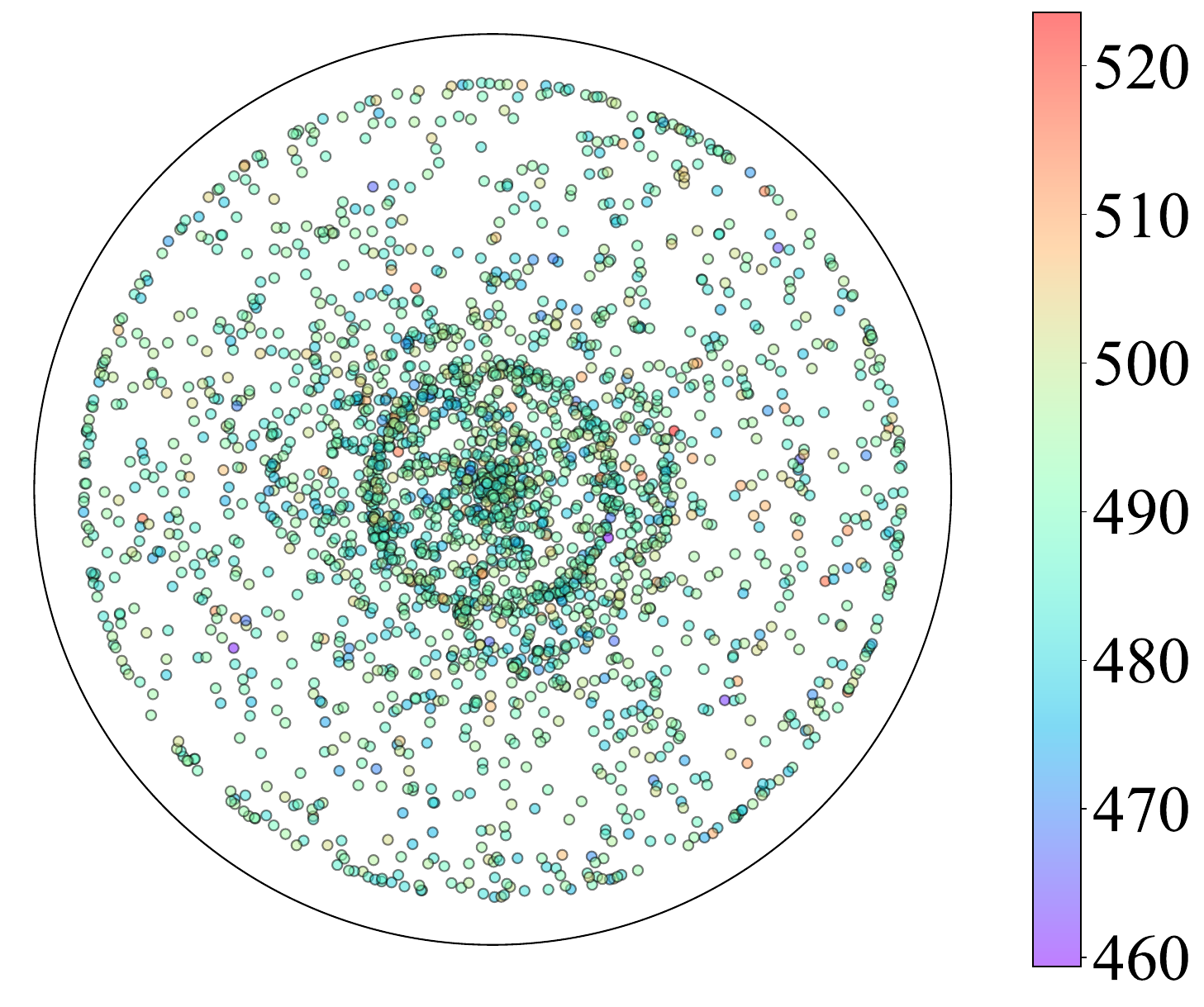}
}%
\centering
\caption{Visualization of noise distribution on Poincar\'e disk for LaP and RdDP on Cora. }
\label{fig:visualization_add}
\end{figure}

Fig.~\ref{fig:visualization_add} illustrates the noise distributions of LaP and RdDP, reaffirming once again that perturbations without hierarchical guidance are uniform, insensitive to hierarchical structures, and incapable of addressing the enhanced MIA associated with hierarchy (As the colors and values in the legend show, red represents a greater amount of noise imposed on that node.)

\subsection{Attack Scenario} \label{sec:attack_senario}
We mainly consider the commonly applied \textbf{Membership Inference Attack (MIA)} model on GNNs.
\begin{itemize}[leftmargin=*]
    \item \textbf{Definition (MIA)}: Given an exact input $\mathbf{x}$ and access to the learned model $f(\mathbf{x};\boldsymbol{\theta^{*}})$, an attacker infers whether $\mathbf{x} \in D_{\textrm{train}}$ or not. We model MIA as a \textbf{binary classification task}, where an adversary attempts to determine whether the target node $v$ belongs to the training node-set.
    \item \textbf{Principle}: MIA consists of the target model, shadow model, and attack model. Attackers obtain the inference result from the target model through query operations, guiding the shadow model to mimic the prediction behavior of the target model to train a set of attack data. The attack model is composed of MLPs to execute the attack.
    \item \textbf{Example}: The target model is typically trained in sensitive domains such as medical. When MIA succeeds, there is a significant risk of leaking patients (nodes)' privacy. 
\end{itemize}

\noindent \textbf{MIA Results.} We show the MIA results in Table~\ref{tab:attack} and obtain two conclusions. 
\begin{itemize}[leftmargin=*]
    \item \textbf{Hierarchical information $\mathcal{H}$ enhances the model's inferential capabilities. }
    GCN+$\mathcal{H}$ gets $\mathcal{H}$ as a bias in training, which enhances the attacker's inferential abilities. 
    \item \textbf{PoinDP exhibits outstanding defensive capabilities when facing MIA. }
    In models with protection capabilities, we uniformly set the privacy budget $\epsilon$ to $0.2$ to ensure perturbation fairness. 
    A smaller performance in MIA indicates better model protection. 
\end{itemize}

\noindent\textbf{Analysis of Hyperbolicity on Attack Experiment.}
The \textit{Gromov's $\delta$-hyperbolicity}~\cite{HGCN_ChamiYRL19} value responds to the similarity between a graph and a tree. 
Table~\ref{tab:attack} is evident that Photo exhibits stronger hierarchical characteristics compared to PubMed ($\delta$=0.15$<$1.65), i.e. smaller $\delta$ values imply that the graph is more tree-like (clearer hierarchical structure) and allowing better utilization of the hyperbolic geometry to do, and leads to two conclusions. 
\begin{itemize}[leftmargin=*]
    \item \textbf{The stronger the hierarchical graph structure, the greater the inference capabilities of attackers.} The GCN+$\mathcal{H} $ model, enhanced with hierarchical information, improved the AUC score of the attack by 2.7\% on Photo with a stronger hierarchy, whereas the improvement was only 0.4\% on PubMed with a lower hierarchy. 
    \item \textbf{Euclidean DP methods are less effective in capturing hierarchical structures compared to methods that are sensitive to hyperbolic space. }
    PoinDP reduces attack performance, which means that Photo with higher hierarchicality is better reflected, further demonstrating the effectiveness of hyperbolic geometry in model design.
\end{itemize}

\end{document}